# Stacking-based deep neural network for player scouting in football[1]

Simon Lacan, IMT Nord Europe, France

## Abstract

Datascouting is one of the most known data applications in professional sport, and specifically football. Its objective is to analyze huge database of players in order to detect high potentials that can be then individually considered by human scouts. In this paper, we propose a stacking-based deep learning model to detect high potential football players. Applied on open-source database, our model obtains significantly better results that classical statistical methods.

## I. Introduction and Related Work

Today, data has taken on an important role in sport, particularly football. Expected goals, real-time player statistics, heatmaps are some examples of data applications in this sport. Among all these, one is taking on a more than important role: datascouting. Indeed, many of the world's biggest clubs are now 100% committed to using data in their recruitment processes. Of course, no one communicates about their datascouting methods, as it's far too tricky a subject and a lot of money is often at stake.

In this article, we will present an original approach of datascouting based on AI, more specifically on supervised learning, to help scouts pre-select talent from a large number of players. The idea is to define several different deep neural networks and to combine their output with stacking technic in order to have a global advice. This mimics the process of the head scout that combines the recommendations of various scout agents to have o global better advice. Clearly, no AI can totally replace the eye of the scouts and their experience, but data-driven tools can save them an enormous amount of time, or even give them a referral.

Many work were proposed to identify players' skills from data. [5] and [6] provides extensive overviews of the literature in this domain. For example, [7] proposes a complete decision-making framework to optimize player recruitment. In [8], the authors propose to evaluate individually the players according to their data and then to estimate their integration in a given team with Random Forest algorithm. Even if we also use supervised machine learning algorithm, our approach is very different because our automatic labelling process is based on the evolution of the player in the time and the algorithms we used are stacked deep neural networks which allows accurate identifications of the most promising players. In Section II, we describe the dataset we used. Then, the different deep neural networks are introduced in Section III and the way we combine those through stacking in detailed in Section IV. Last Section concludes.

## II. Description of the dataset

I used the API of API-SPORTS to collect the data of players [1]. It provides a large number of statistics about football, especially for live score applications [2]. My models have been trained on the basis of fifteen statistics provided by API-Football. Here is a list of some statistics used: 'minutes', 'goals', 'assists', 'passes', 'key passes', 'tackles', 'blocks', 'interceptions', 'won duels', 'successful dribbles', 'fouls won', 'duels ratio', 'dribbles ratio', 'fouls committed', 'yellow cards', 'red cards'.

I worked with raw statistics, simply normalized, but also with statistics divided by the number of minutes. This ratio of stats per minutes allows young players who have played few games but of good quality to be highlighted. For the learning phase, I worked with players from leagues all over the world, outside the European top 5. This represents a total of over 7 000 players, which is still a small dataset for a deep learning problem.

As we wanted to use supervised learning, we had to find a way of labeling the players. But how to evaluate a player's potential numerically? It's a tricky

---

[1] Un résumé en français et les applications potentielles sont proposés en appendice de ce papier

question. One idea could be to use the variation in a player's market value over a certain period to assess his evolution. But as you may know, market value does not always mean quality. Indeed, on today's market, players of English, French or Brazilian nationality are sometimes overvalued. A player's market value also varies according to the number of years left on his contract, his hype, the number of jerseys sold and a multitude of other variables that have not real link with his potential. That's why we have chosen another option. We work with the 2020-2022 couple of years. We collect a player's statistics for the 2020/2021 season, then look at which league the player plays in for the 2022/2023 season. If he plays in one of the top 5 leagues, he is labelled at 1. If he plays in one of the top 5-10 European leagues, he is labelled at 0.66. For other European or non-EU top leagues, the label is 0.33. Finally, if the player plays in a second division or has not been transferred to a club in a better class, he is labelled at 0. Figure 1 summarizes how the labeling process works. The interest of this label is that the final network gives us a score between 0 and 1, and the closer the score is to 1, the more promising the player.

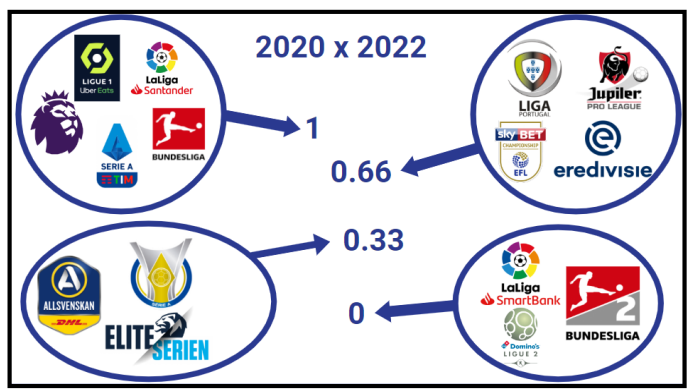

Figure 1: Labeling scheme

Figure 2 shows the distribution of the 4 classes on the dataset. Of course, the most common class is 0, which is explained by the fact that the majority of players are not evolving during 2020-2021 and 2022-2023.

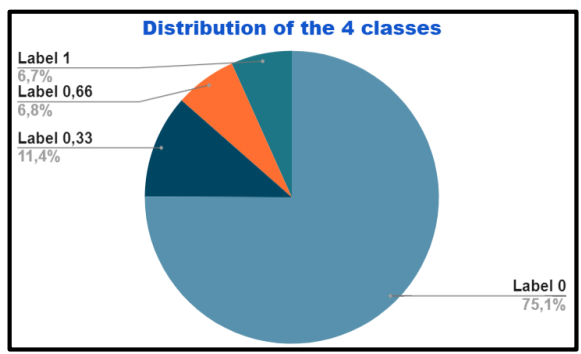

Figure 2: Dataset distribution

## III. First models

I created all my models using the Pytorch framework. To begin with, I trained a neural network on 90% of my dataset, without differentiating the items, then tested it on the remaining 10%. For loss, I used Pytorch's HuberLoss function. For the optimizer, I used Adamw optimizer. Because the classes had a heterogeneous distribution, I used the weighted loss technique to obtain more homogeneous predictions. You can find the formula in appendix 2. The final result of the test phase is the confusion matrix shown in Figure 3. As you can see, the network only manages to make a small difference between players above and below 0.5. So I decided to try training a different network for each position, in the hope that this would increase my results. Figure 4 shows the confusion matrix for the defender test. We can see a progression in the results with increasing accuracy, but it's still not enough if we really want to detect players efficiently.

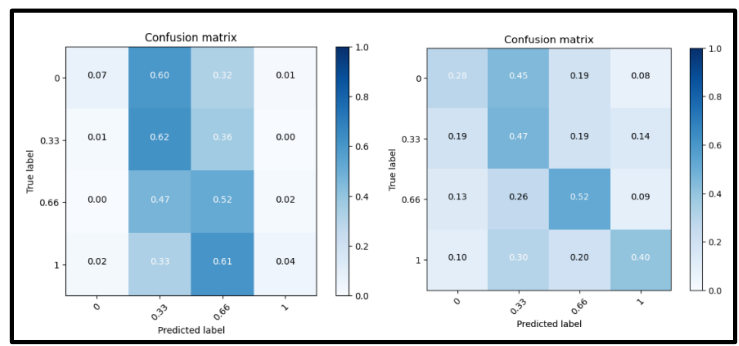

Figure 3 and 4: Confusion matrix of model 1 and 2

## IV. Network stacking and final results

To improve these results, I decided to explore a new intuition. A well-known technique, called network stacking technique takes the outputs of different networks to train another network. So I trained a third network with the outputs of the first two networks introduced in the previous Section. The idea here is to believe that the first two networks make mistakes on different types of players, and that it's possible to create a third model capable of making a better final choice.

For training this model, we don't use weighted loss. The goal is to try to predict a small number of values far from 0, only for interesting players. Figure 5 shows the confusion matrix for the midfielders test.

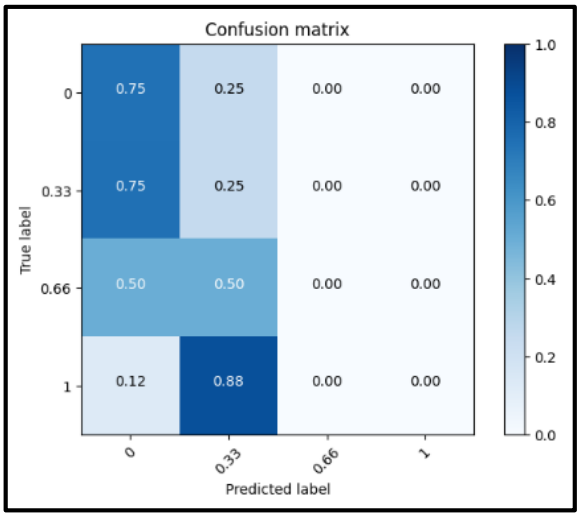

Figure 5: Confusion matrix of stacking model

Not using weighted loss means that all the output values are fairly close to 0. But if we look at the distribution of outputs for each class, we see that the network manages to predict higher values for classes 0.66 and 1. This is confirmed when we calculate the average output class by class.

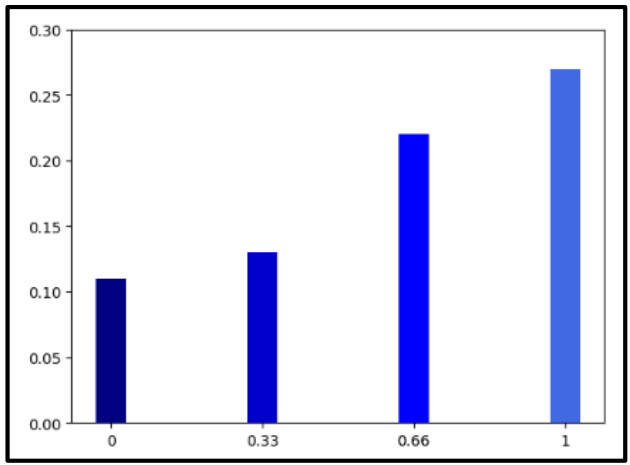

Figure 6: Average output values by classes

If we now assume that we are only interested in players from classes 0.66 and 1, we label this set as 1 and the other players as 0. Next, we define alpha as a threshold value such that the values lower than alpha are labeled 0 and the values higher are labeled 1. If the network predicts an output above alpha, it means that the player interests us. Keeping the predicted outputs from our network and varying alpha between 0 and 0.4, we obtain the following curves on Figure 7. In blue, we have the success rate if we predict 1. If this percentage is, for example, 70%, and we predict 10 players as 1, it means that 7 of our players are truly 1. In orange, we have the percentage of correctly predicted 1s out of the total set of 1s. If this percentage is at 60%, it means that we have predicted as 1 60% of all the players who should have been predicted as 1. We observe that, for example, with alpha = 0.275, we identify 35% of the players to be found, and we make only 17% errors, which is quite acceptable.

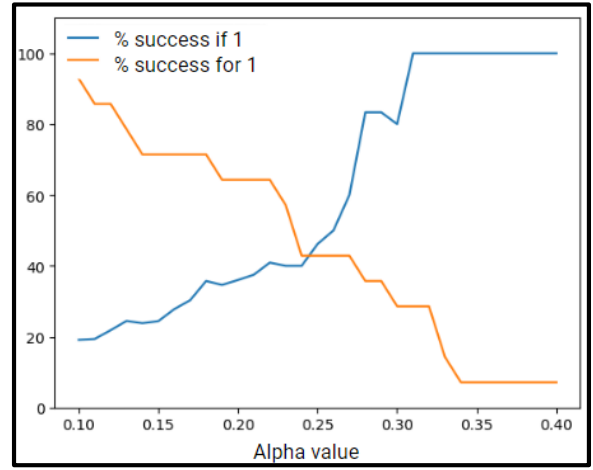

Figure 7: % success based on alpha

With these results, we may fear that above a certain alpha, the network predicts only the players that are easy to find, meaning those with significantly better statistics than others. Therefore, we created a game pitting human football fans against our model. We randomly selected 100 players from our dataset and labeled them. We then asked 3 football fans to rate these players between 0 and 0.4 based on their potential. The participants were not provided with the names or teams of the players, only their statistics. They tried several techniques, such as normalizing the statistics over time, and combined them to obtain an overall score. We then compared the average predictions for each class by the 3 football fans and by our model.

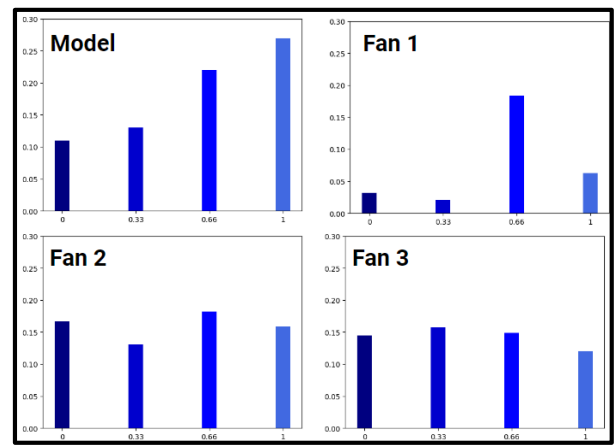

Figure 8: Average output values by classes for the model and the 3 fans

It is noticeable that only our model manages to obtain higher prediction averages for classes 0.66 and 1. Therefore, we can conclude that the results of our model are not trivial, or at least that it is extremely challenging for a human to achieve the same results.

# Conclusion

In conclusion, we can say that our model can be a real aid for scouts. Indeed, it is capable of analyzing the statistics of a large population of players and extracting a small sample in just a few seconds. Obviously, as mentioned in the introduction, nothing can replace the experience of a scout, but this kind of tool can make a difference in this highly competitive field. Note that the stacking technique seems particularly promising. Indeed, by generalizing these results, we plan to use it to combine more than 2 neural networks or even to combine various types of outputs: already AI-based algorithms, human advices, etc. We definitely would be interested by testing these techniques in the scouting context.

Personally, I am aware that this first work is not yet perfect. The choice of labeling is a significant bias, and the small size of the dataset does not aid learning. In few months, at the end of my engineering studies, I would like to apply this technique and more generally develop my skills in this area in a more professional context. That's why I am keenly interested if you have any questions or comments regarding this data scouting work. You can contact me at the following address: simon.lacan@gmail.com.

# References

1: https://rapidapi.com/api-sports/api/api-football

2: https://www.api-football.com/documentation-v3

3: https://soccermatics.readthedocs.io/en/latest/lesson3/PlusMinus.html

4: https://www.kaggle.com/code/emineyetm/football-talent-evaluation-with-machine-learning

5. Ati, A., Bouchet, P., Ben Jeddou, R., Using multi-criteria decision-making and machine learning for football player selection and performance prediction: A systematic review, Data Science and Management (2023)

6. Plakias S, Moustakidis S, Kokkotis C, Papalexi M, Tsatalas T, Giakas G, Tsaopoulos D. Identifying Soccer Players' Playing Styles: A Systematic Review. J Funct Morphol Kinesiol. 2023 Jul 26;8(3):104.

7. Developing A Decision-Making Framework For Player Recruitment In European Football Clubs, Ünsoy, O. (Author). 31 Dec 2022, Student thesis: Phd

8. Ghar, S., Patil, S., & Arunachalam, V. (2021, December). Data Driven football scouting assistance with simulated player performance extrapolation. In 2021 20th IEEE International Conference on Machine Learning and Applications (ICMLA) (pp. 1160-1167). IEEE.

# Appendix

1) Example of a statistics diagram

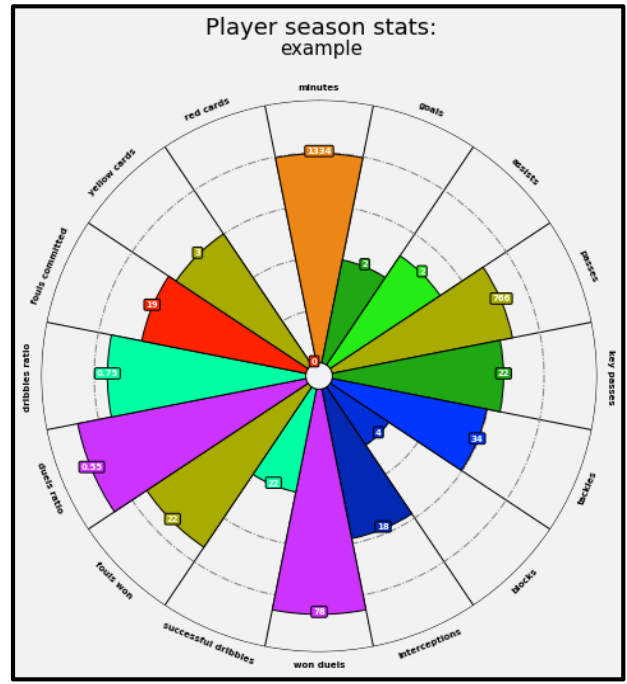

2) Network stacking diagram

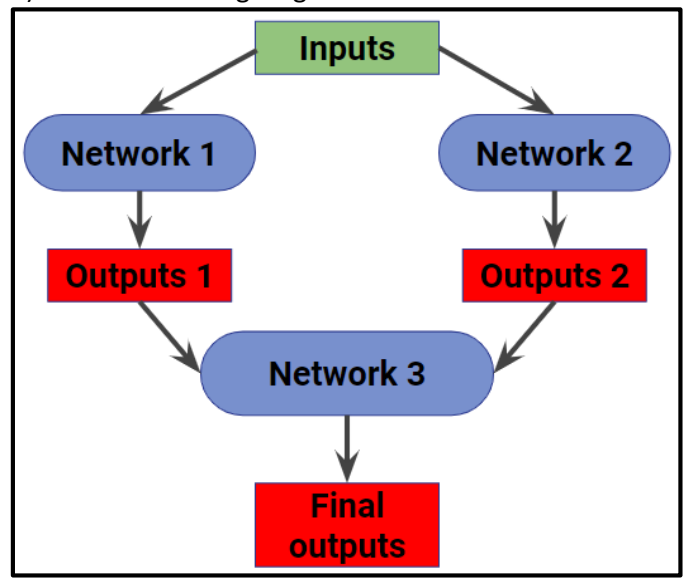

3) Weighted loss formula

$$\sum \frac{Loss(x_k, y_k)}{\mathbb{P}(k)}$$

4) Example of output values for the final model

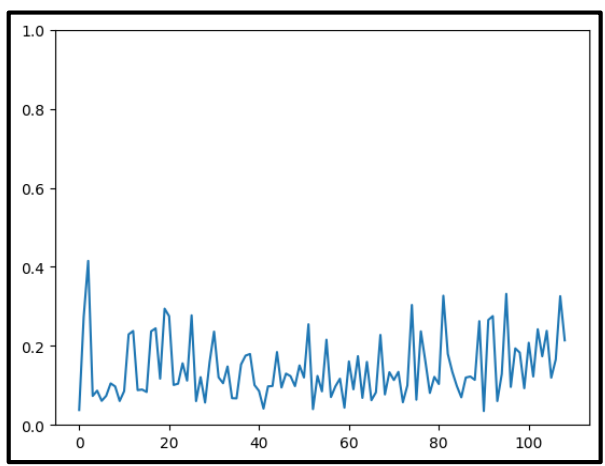

5) Résumé et applications de la méthode proposée

Cet article présente mon premier travail de data appliqué au monde du scouting dans le football. Outre la proposition de labélisation qui peut être qualifié de subjective, l'approche original de celui-ci est l'utilisation du network stacking. Cette technique semble grandement augmenter les prédictions des différents réseaux en combinant plusieurs sorties. L'intérêt est aussi de pouvoir adapter certains sous-réseaux en fonction de plusieurs paramètres comme le poste recherché ou encore le championnat du club. Les limites de ce travail sont essentiellement liées aux données d'entrées qui ne sont pas très diversifiées. L'approche introduite dans cet article semble prometteuse et je souhaite pouvoir la tester dans un contexte plus professionnel.